\documentclass[12pt]{article}  	
\title{Fairness in Criminal Justice Risk Assessments: The State of the Art}
\author{Richard Berk$^{a,b}$, Hoda Heidari$^{c}$,  Shahin Jabbari$^{c}$,  \\ Michael Kearns$^{c}$, Aaron Roth$^{c}$ \\ \\
Department of Statistics$^{a}$ \\
Department of Criminology$^{b}$ \\
Department of Computer and Information Science$^{c}$ \\ \\
University of Pennsylvania}
\begin{document}
\maketitle
\begin{abstract}
\textit{Objectives}: Discussions of fairness in criminal justice risk assessments typically lack conceptual precision. Rhetoric too often substitutes for careful analysis. In this paper, we seek to clarify the tradeoffs between different kinds of fairness and between fairness and accuracy.

\textit{Methods}: We draw on the existing literatures in criminology, computer science and statistics to provide an integrated examination of fairness and accuracy in criminal justice risk assessments. We also provide an empirical illustration using data from arraignments.

\textit{Results}: We show that there are at least six kinds of fairness, some of which are incompatible with one another and with accuracy.

\textit{Conclusions}: Except in trivial cases, it is impossible to maximize accuracy and fairness at the same time, and impossible simultaneously to satisfy all kinds of fairness. In practice, a major complication is different base rates across different legally protected groups. There is a need to consider challenging tradeoffs.
\end{abstract} 
\pagebreak                                      

\section{Introduction}

The use of actuarial risk assessments in criminal justice settings has of late been subject to intense scrutiny.  There have been ongoing discussions about how much better in practice risk assessments derived from machine learning perform compared to risk assessments derived from older, conventional methods (Liu et al., 2011; Berk, 2012; Berk and Bleich, 2013; Brennan and Oliver, 2013; Rhodes, 2013; Ridgeway, 2013a; 2013b). We have learned that when relationships between predictors and the response are complex, machine learning approaches can perform far better. When relationships between predictors and the response are simple, machine learning approaches will perform about the same as conventional procedures. 

Far less close to resolution are concerns about fairness raised by the media (Cohen, 2012; Crawford, 2016; Angwin et al., 2016; Dietrerich et al., 2016; Doleac and Stevenson, 2016), government agencies (National Science and Technology Council, 2016: 30-32), foundations (Pew Center of the States, 2011), and academics (Demuth, 2003; Harcourt, 2007; Berk, 2009; Hyatt et al., 2011; Starr, 2014b; Tonry, 2014; Berk and Hyatt, 2015; Hamilton, 2016).\footnote
{
Many of the issues apply to actuarial methods in general about which concerns have been raised for some time (Berk and Messinger, 1987; Feely and Simon, 1994).
 }
Even when direct indicators of protected group membership, such as race and gender, are not included as predictors, associations between these measures and legitimate predictors can ``bake in'' unfairness. An offender's prior criminal record, for example, can carry forward earlier, unjust treatment not just by criminal justice actors, but by an array of other social institutions that may foster disadvantage. 

As risk assessment critic Sonja Starr writes, ``While well intentioned, this approach [actuarial risk assessment] is misguided. The United States inarguably has a mass-incarceration crisis, but it is poor people and minorities who bear its brunt. Punishment profiling will exacerbate these disparities -- including racial disparities -- because the risk assessments include many race-correlated variables. Profiling sends the toxic message that the state considers certain groups of people dangerous based on their identity. It also confirms the widespread impression that the criminal justice system is rigged against the poor'' (Starr, 2014a).

On normative grounds, such concerns can be broadly legitimate, but without far more conceptual precision, it is difficult to reconcile competing claims and develop appropriate remedies. The debates can become rhetorical exercises, and few minds are changed.

This paper builds on recent developments in computer science and statistics in which fitting procedures, often called algorithms, can assist criminal justice decision-making by addressing both accuracy \textit{and} fairness.\footnote
{
An algorithm is not a model. An algorithm is simply a sequential set of instructions for performing some task. When you balance your checkbook, you are applying an algorithm. A model is an algebraic statement of how the world works. In statistics, often it represents how the data were generated.
}
Accuracy is formally defined by out-of-sample performance using one or more conceptions of prediction error (Hastie et al., 2009: Section 7.2). There is no ambiguity. But, even when attempts are made to clarify what fairness can mean, there are several different kinds that can conflict with one another and with accuracy (Berk, 2016b). 

Examined here are different ways that fairness can be formally defined, how these different kinds of fairness can be incompatible, how risk assessment accuracy can be affected, and various algorithmic remedies that have been proposed. The perspectives represented are found primarily in statistics and computer science because those disciplines are the source of modern risk assessment tools used to inform criminal justice decisions. 

No effort is made here to translate formal definitions of fairness into philosophical or jurisprudential notions in part because the authors of this paper lack the expertise and in part because that multidisciplinary conversation is just beginning (Ferguson, 2015; Barocas and Selbst, 2016; Janssen and Kuk, 2016; Kroll et al., 2017). Nevertheless, an overall conclusion will be that you can't have it all. Rhetoric to the contrary, challenging tradeoffs are required between different kinds of fairness and between fairness and accuracy. 

\section{Confusion Tables, Accuracy, and Fairness}

For ease of exposition and with no important loss of generality, $Y$ is the response variable, henceforth assumed to be binary, and there are two protected group categories: men and women. We begin by introducing by example some key ideas needed later to define fairness and accuracy. We build on the simple structure of a  2 by 2 cross-tabulation (Berk, 2016b; Chouldechova, 2016; Hardt et al., 2016).  Illustrations follow shortly.

\begin{table}[htbp]
\begin{center}
\tiny
\begin{tabular}{|l|c|c|c|}
\hline 
 & Failure Predicted & Success Predicted & Conditional Procedure Error\\
\hline \hline
Failure -- A Positive   &  $a$ & $b$ & $b/(a+b)$ \\
  &   True Positives & False Negatives & False Negative Rate \\ \hline
Success -- A Negative  &  $c$ &  $d$   & $c/(c+d)$ \\
 & False Positives & True Negatives & False Positive Rate \\
\hline 
Conditional Use Error & $c/(a+c)$ & $b/(b+d)$ &  $ \frac{(c+b)}{(a+b+c+d)}$\\
  & Failure Prediction Error & Success Prediction Error & Overall Procedure Error \\
\hline
\end{tabular}
\caption{A Cross-Tabulation of The Actual Outcome by The Predicted Outcome When The Prediction Algorithm Is Applied To A Dataset}
\label{tab:xtab}
\end{center}
\end{table}

Table~\ref{tab:xtab} is a cross-tabulation of the actual binary outcome $Y$ by the predicted binary outcome $\hat{Y}$.  Such tables are in machine learning often called a ``confusion table'' (also ``confusion matrix"). $\hat{Y}$ is the fitted values that result when an algorithmic procedure is applied in the data.  A ``failure'' is called a ``positive" because it motivates the risk assessment; a positive might be an arrest for a violent crime. A ``success'' is a ``negative,'' such as completing a probation sentence without any arrests.  These designations are arbitrary but allow for a less abstract discussion.\footnote
{
Similar reasoning is often used in the biomedical sciences. For example, a success can be a diagnostic test that identifies a lung tumor.
}

The left margin of the table shows the actual outcome classes. The top margin of the table shows the predicted outcome classes. Cell counts internal to the table are denoted by letters. For example, ``$a$'' is the number of observations in the upper-left cell.  All counts in a particular cell have the same observed outcome class and the same predicted outcome class. For example, ``$a$'' is the number of observations for which the observed response class is a failure, and the predicted response class is a failure. It is a true positive. Starting at the upper left cell and moving clockwise around the table are true positives, false negatives, true negatives, and false positives.
 
The cell counts and computed values on the margins of the table can be interpreted as descriptive statistics for the observed values and fitted values in the data on hand. Also common is to interpret the computed values on the margins of the table as estimates of the corresponding probabilities in a population. We turn to that later. 

There is a surprising amount of information that can be extracted from the table. We will use the following going forward.\footnote
{
We proceed in this manner because there will be clear links to fairness. There are many other measures from such a table for which this is far less true. Powers (2011) provides an excellent review. 
}

\begin{enumerate}
\item
\textit{Sample Size} -- The total number of observations conventionally denoted by $N$: $a + b + c + d$.
\item
\textit{Base Rate} -- The proportion of actual failures, which is $(a+b)/(a+b+c+d)$, or the proportion of actual successes, which is $(c+d)/(a+b+c+d)$
\item
\textit{Prediction Distribution} -- The proportion predicted to fail and the proportion predicted to succeed: $(a+c)/(a+b+c+d)$ and $(b+d)/(a+b+c+d)$ respectively.
\item
\textit{Overall Procedure Error} -- The proportion of cases misclassified: $(b+c)/(a+b+c+d)$.
 \item
\textit{Conditional Procedure Error} -- The proportion of cases  incorrectly classified conditional on one of the two \textit{actual} outcomes: $b/(a+b)$, which is the \textit{false negative rate}, and $c/(c+d)$, which is the \textit{false positive rate}.
\item
\textit{Conditional Use Error} -- The proportion of cases incorrectly predicted conditional on one of the two \textit{predicted} outcomes: $c/(a+c)$, which is the proportion of incorrect failure predictions, and  $b/(b+d)$, which is the proportion of incorrect success predictions.\footnote
{
 There seems to be less naming consistency for these kinds errors compared to false negatives and false positives.  Discussions in statistics about generalization error (Hastie et al., 2009: Section 7.2), can provide one set of terms whereas concerns about errors from statistical tests can provide another. In neither case, moreover, is the application to confusion tables necessarily natural. Terms like the ``false discover rate''  and the ``false omission rate,'' or  ``Type II" and ``Type I'' errors can be instructive for interpreting statistical tests but build in content that is not relevant for prediction errors. There is no null hypothesis being tested. 
 }
We use the term conditional \textit{use} error because when risk is actually determined, the predicted outcome is employed; this is how risk assessments are used in the field. 
\item
\textit{Cost Ratio} -- the ratio of false negatives to false positives $b/c$ or the ratio of false positives to false negatives $c/b$.
\end{enumerate}

The discussion of fairness to follow uses all of these features of Table~\ref{tab:xtab}, although the particular features employed will vary with the kind of fairness. We will see, in addition, that the different kinds of fairness can be related to one another and to accuracy. But before getting into a more formal discussion, some common fairness issues will be illustrated with three hypothetical confusion tables.

\begin{table}[htp]
\caption{FEMALES: FAIL(\textit{f}) OR SUCCEED(\textit{s}) ON PAROLE  (Success Base Rate = 500/1000 = .50, Cost ratio = 200/200 = 1:1, Predicted to Succeed 500/1000 = .50)}
\begin{center}
\tiny
\begin{tabular}{|c|c|c|c|}
\hline 
      & $\hat{Y}_{f}$ & $\hat{Y}_{s}$ & Conditional Procedure Error  \\ \hline
 $Y_{f}$ -- Positive  & 300 & \textit{200} & .40 \\
      & True Positives & False Negatives & False  Negative Rate\\   \hline 
 $Y_{s}$  -- Negative & \textit{200} & 300 & .40 \\
      & False Positives & True Negatives & False Positive Rate  \\  \hline
 Conditional Use Error & .40 & .40 & \\ 
      & Failure Prediction Error & Success Prediction error & \\
 \hline
\end{tabular}
\end{center}
\label{tab:females}
\end{table}

Table~\ref{tab:females} is a confusion table for a hypothetical set of women released on parole. Gender is the protected individual attribute.  As failure on parole is a ``positive," and a success on parole is a ``negative." For ease of exposition, the counts are meant to produce a very simple set of results. 

The base rate for success is .50 because half of the women are not re-arrested. The algorithm correctly predicts that the proportion who succeed on parole is .50. This is a favorable initial
 indication of the algorithm's performance because the marginal distribution of $Y$ and $\hat{Y}$ is the same. 

The false negative rate and false positive rate of .40 is the same for successes and failures. When the outcome is known, the algorithm can correctly identify it 60\% of the time. The cost ratio is, therefore, 1 to 1.

The prediction error of .40 is the same for predicted successes  and predicted failures.
When the outcome is predicted, the prediction is correct 60\% of the time.   There is no consideration of fairness because Table~\ref{tab:females} shows only the results for women.

\maketitle
\begin{table}[htp]
\caption{MALES: FAIL(\textit{f}) OR SUCCEED(\textit{s}) ON PAROLE  (Success Base Rate = 500/1500 = .33, Cost ratio 400/200 = 2:1, Predicted to Succeed 700/1500 = .47)}
\begin{center}
\tiny
\begin{tabular}{|c|c|c|c|}
\hline
      & $\hat{Y}_{f}$ & $\hat{Y}_{s}$ &  Conditional Procedure Error  \\ \hline
 $Y_{f}$ -- Positive  & 600 & 400 & .40 \\
      & True Positives & False Negatives & False  Negative Rate\\   \hline 
 $Y_{s}$  -- Negatives & 200 & 300 & .40 \\
      & False Positives & True Negative & False Positive Rate  \\  \hline
     Conditional Use Error & .25 & .57 & \\ 
      & Failure Prediction Error & Success Prediction error & \\
 \hline
\end{tabular}
\end{center}
\label{tab:males}
\end{table}

Table~\ref{tab:males} is a confusion table for a hypothetical set of men released on parole. To help illustrate fairness concerns, the base rate for success on parole is changed from .50 to .33. Men are substantially less likely to succeed on parole than women. The base rate was changed by multiplying the top row of cell counts in Table~\ref{tab:females} by 2.0. That is the \textit{only} change made to the cell counts. The bottom row of cell counts are unchanged.  

The false negative and false positive rates are the same and unchanged at .40. Just as for women, when the outcome is known, the algorithm can correctly identify it 60\% of the time. We will see later that the important comparison is across the two tables. Having a false positive and false negative rate within a table the same, does not figure in definitions of fairness. What matters is whether the false negative rate varies across tables and whether the false positive rate varies across tables. 

Failure prediction error is reduced from .40 to .25, and success prediction error is increased from .40 to .57. Men are more often predicted to succeed on parole when they actually don't. Women are more often predicted to fail on parole when they don't. If predictions of success on parole make a release more likely, some would argue that the prediction errors unfairly favor men. Some would assert more generally that different prediction error proportions for men and women is by itself unfair. 

Whereas in Table~\ref{tab:females}, .50 of the women are predicted to succeed, in Table~\ref{tab:males}, .47 of the men are predicted to succeed. This is a small difference in practice, but it favors women. Some would call this unfair, but it is a different kind of unfairness than disparate prediction errors by gender. 

Perhaps more important, although the proportion of women predicted to succeed corresponds to the actual proportion of women who succeed, the proportion of men predicted to succeed is a substantial overestimate of the actual proportion of men who succeed. For men, the distribution $Y$ is not the same as the distribution of $\hat{Y}$. Some might argue that this makes the algorithmic results overall less defensible for men because a kind of accuracy has been sacrificed. (One would arrive at the same conclusion using predictions of failure on parole). Fairness issues could arise if decision-makers, noting the disparity between the actual proportion who succeed on parole and the predicted proportion who succeed on parole, discount the predictions for men. Predictions of success on parole would be taken less seriously for men than women. 

Finally, the cost ratio in Table~\ref{tab:females} for women makes false positives and false negatives equally costly (1 to 1). In Table~\ref{tab:males}, false positives are twice as costly as false negatives. Incorrectly classifying a success on parole as failure is twice as costly for men (2 to 1). This too can be seen as unfair. Put another way, individuals who succeed on parole but who would be predicted to fail, are of greater relative concern when the individual is a man.    

Note, that all of these potential unfairness and accuracy problems surface solely by changing the base rate even when the false negative rate and false positive rates are unaffected. Base rates can matter a great deal, which is a theme to which we will return.

\begin{table}[htp]
\caption{MALES TUNED: FAIL(\textit{f}) OR SUCCEED(\textit{s}) ON PAROLE  (Success Base Rate = 500/1500 = .33, Cost ratio = 200/200 = 1:1, Predicted to succeed 500/1500 = .33)}
\begin{center}
\tiny
\begin{tabular}{|c|c|c|c|}
\hline
      & $\hat{Y}_{f}$ & $\hat{Y}_{s}$ & Conditional Procedure Error  \\ \hline
 $Y_{f}$ -- Positive  & 800 & \textit{200} & .20 \\
      & True Positives & False Negatives & False  Negative Rate\\   \hline 
 $Y_{s}$  -- Negative & \textit{200} & 300 & .40 \\
      & False Positives & True Negatives & False Positive Rate  \\  \hline
     Conditional Use Error & .20 & .40 & \\ 
      & Failure Prediction Error & Success Prediction error & \\
 \hline
\end{tabular}
\end{center}
\label{tab:tunedmales}
\end{table}

We will see later that there are a number of proposals that try to correct for various kinds of unfairness, including those illustrated in the comparisons between Table~\ref{tab:females} and Table~\ref{tab:males}. For example, it is sometimes possible to tune classification procedures to reduce or even eliminate some forms of unfairness. 

In Table~\ref{tab:tunedmales}, the success base rate for men is still .33, but the cost ratio for men is again 1 to 1. Now, when success on parole is predicted, it is incorrect 40 times out of 100 and corresponds to .40 success prediction error for women. When predicting success on parole, we have equal accuracy for men and women. One kind of unfairness has been eliminated. Moreover, the fraction of men predicted to succeed on parole now equals the actual fraction of men who succeed on parole. Some measure of credibility has been restored to the predictions for men. 

However, the false negative rate for men is now .20, not .40, as it is for women. In trade, therefore, when men actually fail on parole, the algorithm is more likely than for women to correctly identify it. By this measure, the algorithm performs better for men.  Tradeoffs of these kinds are endemic in classification procedures that try to correct for unfairness. Some tradeoffs are inevitable and some are simply common. This too is a theme to which we will return. 

\section{The Statistical Framework}

We have considered confusion tables as descriptive tools for data on hand. The calculations on the margins of the table are proportions. Yet, those proportions are often characterized as probabilities. Implicit are properties that cannot be deduced from the data alone. Commonly, reference to a data generation process is required (Berk, 2016: 11 -- 27; Kleinberg at al., 2016). For clarity, we need to consider that data generation process.

There are practical concerns as well requiring a ``generative'' formulation. In many situations, one wants to draw inferences beyond the data being analyzed. Then, the proportions can be seen as statistical estimates. For example, a confusion table for release decisions at arraignments from a given month, might be used to draw inferences about a full year of arraignments in that jurisdiction (Berk et al., 2016). Likewise, a confusion table for the housing decisions made for prison inmates (e.g., low security housing versus high security housing) from a given prison in a particular jurisdiction might be used to draw inferences about placement decisions in other prisons in the same jurisdiction (Berk and de Leeuw, 1999).\footnote
{
The binary response might be whether an inmate is reported for serious misconduct such as an assault on a guard or another inmate.
}
But perhaps most important, algorithmic results from a given dataset are commonly used to inform decisions in the future. Generalizations are needed over time. 

In such circumstances, one needs a formal rationale for how the data came to be and for the estimation target. In conventional survey sample terms, one must specify a population and one or more population parameters whose values are to be estimated from the data. Probability sampling then provides the requisite justification for statistical inference.

There is a broader formulation that is usually more appropriate for algorithmic procedures. The formulation has each observation randomly realized from a single joint probability distribution. This is a common approach in computer science, especially for machine learning (Kearns, 1994: Section 1.2; Bishop, 2006: Section 1.5), and also can be found in econometrics (White, 1980) and statistics (Freedman, 1981; Buja et al., 2017). 

In this paper, we denote that joint probability distribution by $P(Y,L,S)$. $Y$ is the outcome of interest. An arrest while on probation is an illustration. $L$ includes ``legitimate'' predictors such as prior convictions. $S$ includes ``protected'' predictors such as race, ethnicity and gender. In computer science, $P(Y,L,S)$ often is called a ``target population.''

$P(Y,L,S)$ has all of the usual moments. From this, the population can be viewed as the limitless number of observations that could be realized from the joint probability distribution (as if by random sampling), each observation an IID realized case. Under this conception of a population, all moments and conditional moments are necessarily expectations.

There is in the population some true function of $L$ and $S$, $f(L,S)$, linking the predictors to the expectations of $Y$: $E(Y|L,S)$. When $Y$ is categorical, these conditional expectations are conditional probabilities. $E(Y|L,S)$ is the ``true response surface.'' The data on hand are a set of IID realized observations from $P(Y,L,S)$. In some branches of computer science, such as machine learning, each realized observation is called an ``example.'' 

A fitting procedure, $h(L,S)$, is applied to the data that contain a response $Y$. The \textit{structure} of fitting procedure $h(L,S)$ could be a linear regression model. The optimization \textit{algorithm} could be minimizing the sum of the squared residuals by solving the normal equations. The fitted values after optimization are the estimates. These concepts apply to more flexible fitting procedures as well. For example, the structure of the machine learning procedure gradient boosting is regression trees. The optimization algorithm is gradient descent. 

We allow $S$ to participate in the fitting procedure because it is associated with $Y$. The wisdom of proceeding in this manner is considered later when we address exactly what is being estimated. We denote the fitted values by $\hat{Y}$. The algorithmically produced $\hat{f}(L,S)$, which is the source of $\hat{Y}$, is in computer science an ``hypothesis.''

For all of the usual reasons, $\hat{f}(L,S)$ will almost certainly be a biased estimate of the true response surface.\footnote
{
Formally, a richer notational scheme should be introduced at this point, but we hope the discussion is sufficiently clear without the clutter that would follow. See Buja and his colleagues (2017) for a far more rigorous treatment using proper notation.
} 
For example, some important legitimate predictions may not be available or are measured with error, and there is no guarantee whatsoever that any functional form arrived at for $\hat{f}(L,S)$ is correct. Indeed, for a variety of technical reasons, the algorithms themselves will rarely provide even asymptotically unbiased estimates. For example, procedures that rely on ensembles of decision trees typically fit those trees with ``greedy'' algorithms that can make the calculations tractable at a cost of fitting asymptotically biased approximations of the true response surface.\footnote
{
Consider a single decision tree. Rather than trying all possible trees and picking the tree with the best performance, the fitting proceeds in a sequential stagewise fashion. As the tree is grown, earlier branches are not reconsidered as later branches are determined (Hastie et al., 2009: Section 9.2). 
}
This will invalidate conventional statistical tests and confidence intervals. There are ways to reformulate the inference problem that when coupled with resampling procedures can lead to proper statistical tests and confidence intervals, which are briefly discussed later.  

Defining a population through a joint probability distribution may strike some readers as odd. But if one is to extend empirical results beyond the data on hand, any generalizations must be to something. Considerations of accuracy and fairness require such extensions because a $\hat{f}(L,S)$ developed on a given dataset will be used with new observations that it has not seen before. 
 
A joint probability distribution is essentially an abstraction of a high-dimensional histogram from a finite population. It is just that the number of observations is now limitless, and there is no binning.\footnote
{
As a formal matter, when all of the variables are continuous, the proper term is a joint density because densities rather than probabilities are represented. When the variables are all discrete, the proper term is a joint probability distribution because probabilities are represented. When one does not want to commit to either or when some variables are continuous and some are discrete, one commonly uses the term joint probability distribution. That is how we proceed here. 
}
We imagine that the data are realized by the equivalent of random sampling. We say that the data are realized independently from the same distribution, sometimes denoted by IID for independently and identically distributed. Proper estimation can follow, but now with inferences drawn to an infinite population, or with the same reasoning, to the joint probability distribution that characterizes it. 

Whether this perspective on estimation makes sense for real data depends on substantive knowledge and knowledge about how the data were actually produced. For example, one might be able to make the case that for a particular jurisdiction, all felons convicted in a given year can usefully be seen as IID realizations from all convicted felons that could have been produced that year and perhaps for a few years before and a few years after. One would need to argue that for the given year, or proximate years, there were no meaningful changes in any governing statutes, the composition of the sitting judges, the mix of felons, and the practices of police, prosecutors, and defense attorneys. A more detailed consideration would for this paper be a distraction, and is discussed elsewhere in an accessible, linear regression setting (Berk et al., 2017).

\section{Defining Fairness}

 \subsection{Definitions of Algorithmic Fairness}
 
 We are now ready to consider definitions of algorithmic fairness. Instructive definitions can be found in computer science (Pedreschi, 2008; Kamishima et al., 2011; Dwork et al., 2012; Kamiran et al., 2012; Chouldechova, 2016; Friedler et al., 2016; Hardt et al., 2016;  Joseph et al., 2016; Kleinberg et al., 2016; Calmon et al., 2017; Corbitt-Davies et al., 2017), criminology (Berk, 2016b; Angwin et al., 2016; Dieterich et al., 2016) and statistics (Johnson et al., 2016; Johndrow and Lum, K. 2017). All are recent and focused on algorithms used to inform real-world decisions in both public and private organizations.
 
All of the definitions are broadly similar in intent. What matters is the treatment of protected groups. But the definitions can differ in substantive and technical details. There can be frustrating variation in notation combined with subtle differences in how key concepts are operationalized. There also can be a conflation of information provided by an algorithm and decisions that can follow.\footnote
{
The meaning of ``decision'' can vary. For some it is assigning an outcome class to a numeric risk score. For others, it is an concrete action taken with the information provided by a risk assessment. 
}
We focus here on algorithms. How those can affect decisions is a very important, but different matter (Berk, 2017; Kleinberg et al., 2017). 

Our exposition is agnostic with respect to how outcome classes are assigned by an algorithm and about the fitting procedure used. This is in the spirit of work by Kleinberg and his colleagues (2016). The notation is arithmetically based to facilitate accessibility. 
 
In order to provide clear definitions of algorithmic fairness, we will proceed for now as if $\hat{f}(L,S)$ provides estimates that are the same as the corresponding population features. In this way, we do not conflate a discussion of fairness with a discussion of estimation accuracy. The estimation accuracy is addressed later. We draw heavily on our earlier discussion of confusion tables, but to be consistent with the fairness literature, we emphasize accuracy rather than error. Nevertheless, the notation is drawn from Table~\ref{tab:xtab}. Finally, in applications, there will be a separate confusion table for each class in the protected group. Comparisons are made between these tables. 

\begin{enumerate}
\item
\textit{Overall accuracy equality} is achieved by $\hat{f}(L,S)$ when overall procedure accuracy is the same for each protected group category (e.g., men and women). That is, $(a+d)/(a+b+c+d)$ should be the same (Berk, 2016b). This definition assumes that true negatives are as desirable as true positives. In many settings they are not, and a cost-weighted approach is required. For example, true positives may be twice as desirable as true negatives. Or put another way, false negatives may be two times more undesirable than false positives. Overall accuracy equality is not commonly used because it does not distinguish between accuracy for successes and accuracy for failures. Nevertheless, it has been mentioned in some media accounts (Angwin et al., 2016), and is related in spirit to ``accuracy equity'' as used by Dieterich and colleagues (2016).\footnote
{
Dieterich and his colleagues (2016: 7) argue that overall there is accuracy equity because ``the AUCs obtained for the risk scales were the same, and thus equitable, for blacks and whites.'' The AUC depends on the true positive rate and false positive rate, which condition on the known outcomes. Consequently, it differs formally from overall accuracy equality. Moreover, there are alterations of the AUC that can lead to more desirable performance measures (Powers, 2011). 
}
\item
\textit{Statistical parity} is achieved by $\hat{f}(L,S)$ when the marginal distributions of the predicted classes are the same for both protected group categories. That is, $(a+c)/(a+b+c+d)$ and  $(b+d)/(a+b+c+d)$, although typically different from one another, are the same for both groups (Berk, 2016b).  For example, the proportion of inmates predicted to succeed on parole should be the same for men and women parolees. When this holds, it also holds for predictions of failure on parole because the outcome is binary. This definition of statistical parity, sometimes called ``demographic parity,'' has been properly criticized because it can lead to highly undesirable decisions (Dwork et al., 2012). One might incarcerate women who pose no public safety risk so that the same proportions of men and women are released on probation. 
\item
\textit{Conditional procedure accuracy equality} is achieved by $\hat{f}(L,S)$ when conditional procedure accuracy is the same for both protected group categories (Berk, 2016b). In our notation, $a/(a+b)$ is the same for men and women, and $d/(c+d)$ is the same for men and women. Conditioning on the known outcome, is $\hat{f}(L,S)$ equally accurate across protected group categories? This is the same as considering whether the false negative rate and the false positive rate, respectively, is the same for men and women. Conditional procedure accuracy equality is a common concern in criminal justice applications (Dieterich et al., 2016). Hardt and his colleagues (2016: 2-3) use the term ``equalized odds'' for a closely related definition, and there is a special case they call ``equality of opportunity'' that effectively is the same as our conditional procedure accuracy equality, but only for the outcome class that is more desirable.\footnote
{
One of the two outcome classes is deemed more desirable, and that is the outcome class for which there is conditional procedure accuracy equality. In criminal justice settings, it can be unclear which outcome class is more desirable. Is an arrest for burglary more or less desirable than an arrest for a straw purchase of a firearm? But if one outcome class is recidivism and the other outcome class is no recidivism, equality of opportunity refers to conditional procedure accuracy equality for those who did not recidivate. 
}
\item
\textit{Conditional use accuracy equality} is achieved by $\hat{f}(L,S)$ when conditional use accuracy is the same for both protected group categories (Berk., 2016b).  One is conditioning on the algorithm's \textit{predicted} outcome not the actual outcome. That is, $a/(a+c)$ is the same for men and women, and $d/(b+d)$ is the same for men and women.  Conditional use accuracy equality has also been a common concern in criminal justice risk assessments (Dieterich et al., 2016). Conditional on the prediction of success (or failure), is the probability of success (or failure) the same across groups? Kleinberg and colleagues (2016: 2-4) have a closely related definition that builds on ``calibration.'' More will be said about calibration later. Chouldechova (2016:  section 2.1) arrives at a definition that is the same as conditional use accuracy equality but also only for the outcome class labeled ``positive.'' Her ``positive predictive value'' corresponds to our $a/(a+c)$.\footnote
{
As noted earlier, ``positive" refers to the outcome class that motivates the classification exercise. That outcome class does not have to desirable. We have been calling the outcome class recidivism ``positive.'' 
} 
\item
\textit{Treatment equality} is achieved by $\hat{f}(L,S)$ when the ratio of false negatives and false positives (i.e., $c/b$ or $b/c$) is the same for both protected group categories. The term ``treatment'' is used to convey that such ratios can be a policy lever with which to achieve other kinds of fairness. For example, if false negatives are taken to be more costly for men than women so that conditional procedure accuracy equality can be achieved, men and women are being treated differently by the algorithm. Incorrectly classifying a failure on parole as a success, say, is a bigger mistake for men. The relative numbers of false negatives and false positives across protected group categories also can by itself be viewed as a problem in criminal justice risk assessments (Angwin et al., 2016). Chouldechova (2016: section 2.1) addresses similar issues, but through the false negative rate and the false positive rate: our $b/(a+b)$ and $c/(c+d)$, respectively. 
\item
\textit{Total fairness} is achieved  by $\hat{f}(L,S)$ when (1) overall accuracy equality, (2) statistical parity, (3) conditional procedure accuracy equality, (4) conditional use accuracy equality, and (5) treatment equality are all achieved. Although a difficult pill for some to swallow, we will see that in practice, total fairness cannot be achieved.
\end{enumerate}

Each of the definitions of fairness apply when there are more than two outcome categories. However, there are more statistical summaries that need to be reviewed. For example, when there are three response classes, there are three ratios of false negatives to false positives to be examined. There are also other definitions of fairness not discussed because they currently cannot be operationalized in a useful manner. For example, nearest neighbor parity is achieved if similarly situated individuals are treated similarly (Dwork et al., 2012). Similarly situated is measured by the Euclidian distance between the individuals in predictor space. Unfortunately, the units in which the predictors are measured can make an important difference, and standardizing them just papers over the problem. This is a well-known difficulty with all nearest neighbor methods (Hastie et al., 2009: Chapter 13).

\section{Estimation Accuracy}

We build now on work by Buja and his colleagues (2017). When the procedure $h(L,S)$ is applied to the IID data, the resulting $\hat{f}(L,S)$ can be seen as estimating the true response surface. But even asymptotically, there is no credible claim that the true response surface is being estimated in an unbiased manner. The same applies to the probabilities from a cross-tabulation of $Y$ by $\hat{Y}$. With larger samples, the random estimation error is smaller. On the average, the estimates are closer to the truth. However, the gap between the estimates and the truth combines bias and variance. That gap is not a conventional confidence interval, nor can it be transformed into one. One would have to remove the bias, and to remove the bias, one would need to compare the estimates to the truth. But, the truth is unknown.\footnote
{
If the truth were known, there would be need for the estimates.
}

Alternatively, $\hat{f}(L,S)$ can also be seen estimating a response surface in the population that is an acknowledged \textit{approximation} of the true response surface. In the population, the approximation has the same form as $\hat{f}(L,S)$. Therefore, the estimates of probabilities from Table~\ref{tab:xtab} can be estimates of the corresponding probabilities from a $Y$ by $\hat{Y}$ cross-tabulation if $h(L,S)$ were applied in the population. Thanks to the IID nature of the data, these estimates can also be asymptotically unbiased so that in large samples, the bias will likely be relatively small. This allows one to use sample results to address fairness as long as one appreciates that it is fairness measured by the approximation, not the true $E(Y|L,S)$. Because it is the performance of $\hat{f}(L,S)$ that matters, a focus on $\hat{f}(L,S)$ is consistent with policy applications.\footnote
{
There are technical details that are beyond the scope of this paper. Among the key issues is how the algorithm is tuned and the availability of test data as well as training data (Berk, 2016a). There are also disciplinary differences in how important it is to explicitly estimate features of the approximation. For many statisticians, having an approximation as an estimation target is desirable because one can obtain asymptotically the usual array of inferential statistics. For many computer scientists, such statistics are of little interest because what really matters is estimates of the true response surface.
}

For either estimation target, estimation accuracy is addressed by out-sample-performance. Fitted values in-sample will be subject to overfitting. In practice, this means using test data, or some good approximation thereof, with measures of fit such as generalization error or expected prediction error (Hastie et al., 2006: Section 7.2). Often, good estimates of accuracy may be obtained, but the issues can be tricky. Depending on the procedure $h(L,S)$ and the availability of an instructive form of test data, there are different tools that vary in their assumptions and feasibility (Berk, 2016a). With our focus on fairness, such details are a diversion. 

In summary, $h(L,S)$ can best be seen as a procedure to approximate the true response surface. If the estimation target is the true response surface, the estimates will be biased although random estimation error will be smaller in larger samples. It follows that the probabilities estimated from a cross-tabulation of $Y$ by $\hat{Y}$ will have the same strengths and weaknesses. If the estimation target is the acknowledged approximation, the approximate response surface can be estimated in an asymptotically unbiased manner with less random estimation error the larger the sample size. This holds for the probabilities computed from a cross-tabulation of $Y$ by $\hat{Y}$. Under either estimation target, proper measures of accuracy often can be obtained with test data or methods that approximate test data such as the nonparametric bootstrap (Hastie et al., 2009: Section 8.2).

\section{Tradeoffs}

We turn to tradeoffs and begin by emphasizing an obvious point that can get lost in discussions of fairness. If the goal of applying $h(L,S)$ is to capitalize on non-redundant associations that $L$ and $S$ have with the outcome, excluding $S$ will reduce accuracy. Any procedure that even just discounts the role of $S$ will lead to less accuracy. The result is a larger number of false negatives and false positives. For example, if $h(L)$ is meant to help inform parole release decisions, there will likely be an increase in both the number of inmates who are unnecessarily detained and the number of inmates who are inappropriately released. The former victimizes inmates and their families. The latter increases the number of crime victims. But fairness counts too, so we need to examine tradeoffs.  

Because the different kinds of fairness defined earlier share cell counts from the cross-tabulation of $Y$ against $\hat{Y}$, and because there are relationships between the cell counts themselves (e.g., they sum to the total number of observations), the different kinds of fairness are related as well. It should not be surprising, therefore, that there can be tradeoffs between the different kinds of fairness. Arguably, the tradeoff that has gotten the most attention is between conditional use accuracy equality and the false positive and false negative rates (Angwin et al., 2016; Dieterich, 2016; Kleinberg et al., 2016; Chouldechova, 2016). It is also the tradeoff that to date has the most complete mathematical results. 

\subsection{Some Proven ``Impossibility Theorems'' }

We have conveyed informally that there are incompatibilities between different kinds of fairness. It is now time to be specific. We begin with three definitions. They will be phrased in probability terms, but are effectively the same if phrased in terms of proportions. 

\begin{itemize}
\item
\textit{Calibration} -- Suppose an algorithm produces a score that can then be used to assign an outcome class, much in the spirit of the COMPAS instrument. ``Calibration within groups'' requires that for each score value, or for proximate score values, the proportion of people who actually experience a given outcome (e.g., re-arrested in parole) is the same as the proportion of people predicted to experience that outcome.  As noted earlier, this can be taken as an indicator of how well an algorithm performs. But calibration can become a fairness matter if there is calibration within one group but not within the other. There is a lack of ``balance." A decision-maker may be inclined to take the predictions less seriously for the group that lacks calibration (Kleinberg et al., 2016). Even if there is no calibration for either group, different conditional use accuracy has been a salient concern in criminal justice applications and will be emphasized here (Chouldechova, 2016).\footnote
{
There is a bit of definitional ambiguity because Chouldechova (2016: 2) characterizes equal conditional use accuracy as ``well-calibrated.''
}
\item
\textit{Base Rate} -- This too was introduced earlier. In the population, base rates are determined by the marginal distribution of the response. They are the probability of each outcome class. For example, the base rate for succeeding on parole might be .65 and for not succeeding on parole is then .35. If there are $C$ outcome classes, there will be $C$ base rates.

We are concerned here with base rates for different protected group categories, such as men compared to women.  Base rates for each protected group category are said to be equal if they are identical. Is the probability of succeeding on parole .65 for both men and women? 
\item
\textit{Separation} -- In a population, the observations are separable if for each possible configuration of predictor values, there is some $h(L,S)$ for which the probability of membership in a given outcome class is always either 1.0 or 0.0. In other words, perfectly accurate classification is possible. In practice, what matters is whether there is perfect classification when $h(L,S)$ is applied to data. \end{itemize}

And now the impossibility result:  \textit{When the base rates differ by protected group and when there is not separation, one cannot have both conditional use accuracy equality and equality in the false negative and false positive rates} (Kleinberg et al., 2016; Chouldechova, 2016).\footnote
{
This impossibility theorem is formulated a little differently by Kleinberg and his colleagues and by Chouldechova. Kleinberg et al. (2016) impose calibration and make explicit use of a risk scores from the algorithm. There is no formal transition to outcome classes. Chouldechova (2016), does not impose calibration in the same sense, and moves quickly from risk scores to outcome classes. But both sets of results are for our purposes effectively the same and consistent with our statement. 
} 
Put more positively, one needs either equal base rates or separation to achieve at the same time conditional use accuracy equality and equal false positive and false negative rates for each protected group category.

The implications of this impossibility result are huge. First, if there is variation in base rates and no separation, you can't have it all. The goal of complete race or gender neutrality is unachievable. In practice, both requirements are virtually never met, except in highly stylized examples.

Second, altering a risk algorithm to improve matters can lead to difficult stakeholder choices. If it is essential to have conditional use accuracy equality, the algorithm will produce different false positive and false negative rates across the protected group categories. Conversely, if it is essential to have the same rates of false positives and false negatives across protected group categories, the algorithm cannot  produce conditional use accuracy equality. Stakeholders will have to settle for an increase in one for a decrease in the other. To see how all of this can play out, consider the following didactic illustrations.

\subsubsection{Trivial Case \#1: Assigning the Same Outcome Class to All}

Suppose $h(L,S)$ assigns the same outcome class to everyone (e.g., a failure).  Such an assignment procedure would never be used in practice, but it raises some important issues in a simple setting.

Tables~\ref{tab:constantM} and \ref{tab:constantF} provide an example when the base rates are the same for men and women. There are 500 men and 50 women, but the relative representation of men and women does not matter materially in what follows. Failures are coded 1 and successes are coded 0, much as they might be in practice. Each case is assigned failure (i.e., $\hat{Y}=1$), but the same lessons would be learned if each case is assigned a success (i.e., $\hat{Y} = 0$).  A base rate of .80 for failures is imposed on both tables.

\begin{table}[htbp]
\footnotesize
\caption{Males: A Cross-Tabulation When All Cases Are Assigned The Outcome Of Failure (Base Rate $= .80$, N $=500$)}
\begin{center}
\begin{tabular}{|l|c|c|c|}
\hline \hline
Truth & $\hat{Y} = 1$ & $\hat{Y} = 0$ & Conditional Procedure Accuracy  \\
\hline
$ Y=1$ (a positive -- Fail) & 400 & 0  & 1.0 \\
\hline
$Y = 0$  (a negative -- Not Fail)& 100 & 0  & 0.0 \\
\hline
Conditional Use Accuracy & .80 & -- & \\
\hline
\hline
\end{tabular}
\end{center}
\label{tab:constantM}
\end{table}

\begin{table}[htbp]
\footnotesize
\caption{Females: A Cross-Tabulation When All Cases Are Assigned The Outcome of Failure (Base Rate $= .80$, N $=50$)}
\begin{center}
\begin{tabular}{|l|c|c|c|}
\hline \hline
Truth & $\hat{Y} = 1$ & $\hat{Y} = 0$ & Conditional Procedure Accuracy  \\
\hline
$ Y=1$ (a positive -- Fail) & 40 & 0  & 1.0 \\
\hline
$Y = 0$  (a negative -- Not Fail)& 10 & 0  & 0.0 \\
\hline
Conditional Use Accuracy  & .80 & -- & \\
\hline
\hline
\end{tabular}
\end{center}
\label{tab:constantF}
\end{table}

In practice, this approach makes no sense. Predictors are not being exploited. But, one can see that there is conditional procedure accuracy equality, conditional use accuracy equality and overall accuracy equality. The false negative and false positive rates are the same for men and women as well  at 0.0 and 1.0. There is also statistical parity. One does very well on fairness for a risk tool that cannot help decision-makers address risk in a useful manner. Accuracy has been given a very distant backseat. There is a dramatic tradeoff between accuracy and fairness.

If one allows the base rates for men and women differ, there is immediately a fairness price. Suppose in Table~\ref{tab:constantM}, 500 men fail instead of 400. The false positive and false negative rates are unchanged. But because the base rate for men is now larger than the base rate for women (i.e., .83 v. .80), conditional use accuracy is now higher for men, and a lower proportion of men will be incorrectly predicted to fail. This is the sort of result that would likely trigger charges of gender bias. Even in this ``trivial'' case, base rates matter.\footnote
{
When base rates are the same in this example, one perhaps could not achieve perfect fairness while also getting perfect accuracy. The example doesn't have enough information to conclude that the populations aren't separable. But that is not the point we are trying to make.
}

\subsubsection{Trivial Case \#2: Assigning the Classes Using the Same Probability for All}

Suppose each case is assigned to an outcome class with the \textit{same} probability. As in Trivial Case \#1, no use made of predictors, so that accuracy does not figure into the fitting process.

For Tables~\ref{tab:sameM} and \ref{tab:sameF}, the assignment probability for failure is .30 for all, and therefore, the assignment probability for success is .70 for all. Nothing important changes should some other probability be used.\footnote
{
The numbers in each cell assume for arithmetic simplicity that the counts come out exactly as they would in a limitless number of realizations. In practice, an assignment probability of .30 does not require exact cell counts of 30\%.
} 
The base rates for men and women are the same. For both, the proportions that fail are .80.

\begin{table}[htbp]
\footnotesize
\caption{Males: A Cross-Tabulation With Failure Assigned To All With A Probability of .30 (Base Rate $= .80$, N $=500$)}
\begin{center}
\begin{tabular}{|l|c|c|c|}
\hline \hline
Truth & $\hat{Y} = 1$ & $\hat{Y} = 0$ & Conditional Procedure Accuracy  \\
\hline
$ Y=1$ (a positive -- Fail) & 120 & 280  & .30 \\
\hline
$Y = 0$  (a negative -- Not Fail)& 30 & 70  & .70 \\
\hline
Conditional Use Accuracy & .80 & .20 & \\
\hline
\hline
\end{tabular}
\end{center}
\label{tab:sameM}
\end{table}

\begin{table}[htbp]
\footnotesize
\caption{Females: A Cross-Tabulation With Failure Assigned To All With A Probability of .30 (Base Rate $= .80$, N $=50$)}
\begin{center}
\begin{tabular}{|l|c|c|c|}
\hline \hline
Truth & $\hat{Y} = 1$ & $\hat{Y} = 0$ & Conditional Procedure Accuracy  \\
\hline
$ Y=1$ (a positive -- Fail) & 12 & 28  & .30 \\
\hline
$Y = 0$  (a negative -- Not Fail)& 3 & 7  & .70 \\
\hline
Conditional Use Accuracy & .80 & .20 & \\
\hline
\hline
\end{tabular}
\end{center}
\label{tab:sameF}
\end{table}

In Tables~\ref{tab:sameM} and \ref{tab:sameF}, we have the same fairness results we had in Tables~\ref{tab:constantM} and \ref{tab:constantF}, again with accuracy sacrificed. But suppose the second row of entries in Table~\ref{tab:sameF} were 30 and 70 rather than 3 and 7. Now the failure base rate for women is .29, not .80. Conditional procedure accuracy equality remains from which it follows that the false negative and false positive rates are the same as well. But conditional use accuracy equality is lost.  The probabilities of correct predictions for men are again .80 for failures, and .20 for successes. But for women, the corresponding probabilities are .29 and .71. Base rates really matter. 

\subsubsection{Perfect Separation}

We now turn to an $h(L,S)$ that is not trivial, but also very unlikely in practice. In a population, the observations are separable. In Tables~\ref{tab:NoErrorM} and \ref{tab:NoErrorF}, there is perfect separation, and  $h(L,S)$ finds it. Base rates are the same for men and women: .80 fail. 

\begin{table}[htbp]
\footnotesize
\caption{Males: A Cross-Tabulation With Separation and Perfect Prediction (Base Rate $= .80$, N $=500$)}
\begin{center}
\begin{tabular}{|l|c|c|c|}
\hline \hline
Truth & $\hat{Y} = 1$ & $\hat{Y} = 0$ & Conditional Procedure Accuracy  \\
\hline
$ Y=1$ (a positive -- Fail) & 400 & 0  & 1.0 \\
\hline
$Y = 0$  (a negative -- Not Fail)& 0 & 100  & 1.0 \\
\hline
Conditional Use Accuracy & 1.0 & 1.0 & \\
\hline
\hline
\end{tabular}
\end{center}
\label{tab:NoErrorM}
\end{table}

\begin{table}[htbp]
\footnotesize
\caption{Females: A Cross-Tabulation With Separation and Perfect Prediction (Base Rate $=.80$, N $=50$)}
\begin{center}
\begin{tabular}{|l|c|c|c|}
\hline \hline
Truth & $\hat{Y} = 1$ & $\hat{Y} = 0$ & Conditional Procedure Accuracy  \\
\hline
$ Y=1$ (a positive -- Fail) & 40 & 0  & 1.0 \\
\hline
$Y = 0$  (a negative -- Not Fail)& 0 & 10  & 1.0 \\
\hline
Conditional Use Accuracy & 1.0 & 1.0 & \\
\hline
\hline
\end{tabular}
\end{center}
\label{tab:NoErrorF}
\end{table}

There are no false positives or false negatives, so the false positive rate and the false negative rate for both men and women are 0.0. There is conditional procedure accuracy equality and conditional use accuracy equality because conditional procedure accuracy and conditional use accuracy are both perfect. This is the ideal, but fanciful, setting in which we can have it all. 

Suppose for women in Table~\ref{tab:NoErrorF}, there are 20 women who do not fail rather than 10. Their failure base rate for females is now .67 rather than .80. But because of separation, conditional procedure accuracy equality and conditional use accuracy equality remain, and the false positive and false negative rates for men and women are still 0.0. Separation saves the day.\footnote
{
Although statistical parity has not figured in these illustrations, changing the base rate negates it. 
 }
 
\subsubsection{Closer To Real Life}

There will virtually never be separation in the real data even if there there happens to be separation in the joint probability distribution responsible for the data. The fitting procedure $h(L,S)$ may be overmatched because important predictors are not available or because the algorithm arrives at a suboptimal result.  Nevertheless, some types of fairness can sometimes be achieved if base rates are cooperative. 

If the base rates are the same and $h(L,S,)$ finds that, there can be lots of good news. Tables \ref{tab:EqualF} and \ref{tab:EqualM} illustrate. Conditional procedure accuracy equality, conditional use accuracy equality, overall procedure accuracy hold, and the false negative rate and the false positive rate are the same for men and women. Results like those shown in Tables \ref{tab:EqualF} and \ref{tab:EqualM} can occur in real data, but would be rare in criminal justice applications for the common protected groups. Base rates will not be the same.

\begin{table}[htbp]
\footnotesize
\caption{Females: A Cross-Tabulation Without Separation (Base Rate $=.56$, N $=900$)}
\begin{center}
\begin{tabular}{|l|c|c|c|}
\hline \hline
Truth & $\hat{Y} = 1$ & $\hat{Y} = 0$ & Conditional Procedure Accuracy  \\
\hline
$ Y=1$ (a positive -- Fail) & 300 & 200  & .60 \\
\hline
$Y = 0$  (a negative -- Not Fail)& 200 & 200  & .50 \\
\hline
Conditional Use Accuracy & .60 & .50 & \\
\hline
\hline
\end{tabular}
\end{center}
\label{tab:EqualF}
\end{table}

\begin{table}[htbp]
\footnotesize
\caption{Males: Confusion Table Without Separation (Base Rate is $=.56$, N $= 1400$)}
\begin{center}
\begin{tabular}{|l|c|c|c|}
\hline \hline
Truth & $\hat{Y} = 1$ & $\hat{Y} = 0$ & Conditional Procedure Accuracy  \\
\hline
$ Y=1$ (a positive -- Fail) & 600 & 400  & .60 \\
\hline
$Y = 0$  (a negative -- Not Fail) & 400 & 400  & .50 \\
\hline
Conditional Use Accuracy & .60 & .50 & \\
\hline
\hline
\end{tabular}
\end{center}
\label{tab:EqualM}
\end{table}

Suppose there is separation but the base rates are not the same. We are back to Tables~\ref{tab:NoErrorM} and \ref{tab:NoErrorF}, but with a lower base rate. Suppose there is no separation, but the base rates are the same. We are back to Tables~\ref{tab:EqualF} and \ref{tab:EqualM}.

\begin{table}[htbp]
\footnotesize
\caption{Confusion Table For Females With No Separation And A Different Base Rate Compared to Males (Female Base Rate Is $500/900=.56$)}
\begin{center}
\begin{tabular}{|l|c|c|c|}
\hline \hline
Truth & $\hat{Y} = 1$ & $\hat{Y} = 0$ & Conditional Procedure Accuracy  \\
\hline
$ Y=1$ (a positive -- Fail) & 300 & 200  & .60 \\
\hline
$Y = 0$  (a negative -- Not Fail)& 200 & 200  & .50 \\
\hline
Conditional Use Accuracy & .60 & .50 & \\
\hline
\hline
\end{tabular}
\end{center}
\label{tab:NoSepF}
\end{table}

From Tables~\ref{tab:NoSepF} and \ref{tab:NoSepM}, one can see that when there is no separation and different base rates, there can still be conditional procedure accuracy equality. From conditional procedure accuracy equality, the false negative rate and false positive rate, though different from one another, are the same across men and women. This is a start. But treatment equality is gone from which it follows that conditional use accuracy equality has been sacrificed. There is greater conditional use accuracy for women.

\begin{table}[htbp]
\footnotesize
\caption{Confusion Table for Males With No Separation And A different Base Rate Compared to Females (Male Base Rate Is $1000/2200=.45$)}
\begin{center}
\begin{tabular}{|l|c|c|c|}
\hline \hline
Truth & $\hat{Y} = 1$ & $\hat{Y} = 0$ & Conditional Procedure Accuracy  \\
\hline
$ Y=1$ (a positive -- Fail) & 600 & 400  & .60 \\
\hline
$Y = 0$  (a negative -- Not Fail) & 600 & 600  & .50 \\
\hline
Conditional Use Accuracy & .50 & .40 & \\
\hline
\hline
\end{tabular}
\end{center}
\label{tab:NoSepM}
\end{table}

Of the lessons that can be taken from the sets of tables just analyzed, perhaps the most important for policy is that when there is a lack of separation and different base rates across protected group categories, a key tradeoff will be between the false positive and false negative rates on one hand and conditional use accuracy equality on the other. Different base rates across protected group categories would seem to require a thumb on the scale if conditional use accuracy equality is to be achieved. To see if this is true, we now consider corrections that have been proposed to improve algorithmic fairness.
 
\section{Potential Solutions}

There are several recent papers that have proposed ways to reduce and even eliminate certain kinds of bias. As a first approximation, there are three different strategies (Hajian and Domingo-Ferrer, 2013), although they can also be combined when accuracy as well as fairness are considered. 

\subsection{Pre-Processing}

Pre-processing means eliminating any sources of unfairness in the data before $h(L,S)$ is formulated. In particular, there can be legitimate predictors that are related to the classes of a protected group. Those problematic associations can be carried forward by the algorithm. 

One approach is to remove all linear dependence between $L$ and $S$ (Berk, 2009). One can regress in turn each predictor in $L$ on the predictors in $S$, and then work with the their residuals. For example, one can regress predictors such as prior record and current charges on race and gender. From the fitted values, one can construct ``residualized'' transformations of the predictors to be used. 

A major problem with this approach is that interactions effects (e.g., with race and gender) containing information leading to unfairness are not removed unless they are explicitly included in the residualizing regression \textit{even if all of the additive contaminants are removed}. In short, all interactions effects, even higher order ones, would need to anticipated. The approach becomes very challenging if interaction effects are anticipated between $L$ and $S$.

Johndrow and Lum (2017) suggest a far more sophisticated residualizing process that in principle can handle such complications. Fair prediction is defined as constructing fitted values for some outcome using no information from membership in any protected classes. The goal is to transform all predictors so that fair prediction can be obtained ``while still preserving as much `information' in X as possible (Johndrow and Lum, 2017: 6). They formulate this using the Euclidian distance between the original predictors and the transformed predictors. The predictors are placed in order of the complexity of their marginal distribution, and each is residualized in turn using as predictors results from previous residualizations and indicators for the protected class. The regressions responsible for the residualizations are designed to be flexible so that nonlinear relationships can be exploited. But, as Johnson and Lum note, they are only able to consider one form of unfairness. In addition, they risk exacerbating one form of unfairness while mitigating another. 

Base rates that vary over protected group categories can be another source of unfairness. A simple fix is to rebalance the marginal distributions of the response variable so that the base rates for each category are the same. One method is to apply weights for each group separately so that the base rates across categories are the same. For example, women who failed on parole might given more weight, and males who failed on parole might be given less weight. After the weighting, men and women could have a base rate that was the same as the overall base rate. 

A second rebalancing method is to randomly relabel some response values to make the base rates comparable. For example, one could for a random sample of men who failed on parole, recode the response to a successes and for a random sample of women who succeeded on parole, recode the response to a failure. 

Rebalancing has at least two problems. First, there is likely to be a loss in accuracy. Perhaps such a tradeoff between fairness and accuracy will be acceptable to stakeholders, but before such a decision is made, the tradeoff must be made numerically specific. How many more armed robberies, for instance, will go unanticipated in trade for a specified reduction in the disparity between incarceration rates for men and women? Second, rebalancing implies using different false positive to false negative rates for different protected group categories. For example, false positives (e.g., incorrectly predicting that individuals will fail on parole) are treated as relatively more serious errors for men than for women. In addition to the loss in accuracy, stakeholders are trading one kind of unfairness for another.

A third approach capitalizes on association rules, popular in marketing studies (Hastie et al., 2009: section 14.2). Direct discrimination is addressed when features of some protected class are used as predictors (e.g., male). Indirect discrimination is addressed when predictors are used that are related to those protected classes (e.g., arrests for aggravated assault). There can be evidence of either if the conditional probability of the outcome changes when either direct or indirect measures of protected class membership are used as predictors compared to when they are not used. One potential correction can be obtained by perturbing the suspect class membership (Pedreschi et al., 2008). For a random set of cases, one might change the label for man to the label for a woman. Another potential correction can be obtained by perturbing the the outcome label. For a random set of men, one might change failure on parole to success on parole (Hajian and Domingo-Ferrer, 2013). Note that the second approach changes the base rate. We examined earlier the consequences of changing base rates. Several different kinds of fairness can be affected. It can be risky to focus on a single definition of fairness. 

A fourth approach is perhaps the most ambitious. The goal is to randomly transform all predictors except for indicators of protected class membership so that the joint distribution of the predictors is less dependent on protected class membership. An appropriate reduction in dependence is a policy decision. The reduction of dependence is subject to two constraints: (1) the joint distribution of the transformed variables is very close to the joint distribution of the original predictors and (2) no individual cases are substantially distorted because large changes are made in predictor values (Calmon et al., 2017).  An example of a distorted case would be a felon with no prior arrests assigned a predictor value of 20 prior arrests. It is unclear, however, how this procedure maps to different kinds of fairness. For example, the transformation itself may inadvertently treat prior crimes committed by men as less serious than similar prior crimes committed by women -- the transformation may be introducing the prospect of unequal treatment. There are also concerns about the accuracy price, which is not explicitly taken into account. 

\subsection{In-Processing}

In-processing means making fairness adjustments as part of the process by which $h(L,S)$ is constructed. To take a simple example, risk forecasts for particular individuals that have substantial uncertainty can be altered to improve fairness. If whether or not an individual is projected as high risk depends on little more than a coin flip, the forecast of high risk can be changed to low risk to serve some fairness goal. One might even order cases from low certainty to high certainty for the class assigned so that low certainty observations are candidates for alterations first. The reduction in out-of-sample accuracy may well be very small. One can embed this idea in a classification procedure so that explicit tradeoffs are made (Kamiran and Calder, 2009; Kamiran et al., 2016; Corbett-Davies et al. 2017). But this too can have unacceptable consequences for the false positives and false negative rates. A thumb is being put on the scale once again, so there can be inequality of treatment.

A more technically demanding approach is to add a new penalty term to a penalized fitting procedure (Kamishima et al., 2011). Beyond a penalty for an unnecessarily complex fit, there is a penalty for violations of conditional procedure accuracy equality. One important complication is that the loss function typically will not be convex so that local solutions can result. Another important complication is that there will often be undesirable implications for types of unfairness not formally taken into account. 

\subsection{Post-Processing} 

Post-processing means that after $h(L,S)$ is applied, its performance is adjusted to make it more fair. To date, perhaps the best example of this approach draws on the idea of random reassignment of the class label previously assigned by $h(L,S)$ (Hardt et al., 2016). Fairness, called ``equalized odds,'' requires that the fitted outcome classes (e.g., high risk or low risk) are independent of protected class membership, conditioning on the actual outcome classes. The requisite information is obtained from the rows of confusion table and, therefore, represent classification accuracy, not prediction accuracy. There is a more restrictive definition called ``equal opportunity''  requiring such fairness only for the more desirable of the two outcome classes.\footnote
{
In criminal justice applications, determining which outcome is more desirable will often depend on which stakeholders you ask. 
}
 
For a binary response, some cases are assigned a value of 0 and some assigned a value of 1. To each is attached a probability of switching from a 0 to a 1 or from a 1 to a 0 depending in whether a 0 and a 1 is the outcome assigned by $\hat{f}(L,S)$. These probabilities can differ from one another and both can differ across different protected group categories. Then, there is a linear programming approach to minimize the classification errors subject to one of the two fairness constraints. This is accomplished by the values chosen for the various probabilities of reassignment.  The result is an $\hat{f}(L,S)$ that achieves conditional procedure accuracy equality.

The implications of this approach for other kinds of fairness are not clear, and conditional use accuracy (i.e., equally accurate predictions) can be a casualty. It is also not clear how best to build in asymmetric costs of false negatives and false positive. And, there is no doubt that accuracy will decline and will decline more when the probabilities of reassignment are larger. Generally, one would expect to have overall classification accuracy comparable to that achieved for the protected group category for which accuracy is the worst. Moreover, the values chosen for the reassignment probabilities will need to be larger when the base rates across the protected group categories are more disparate. In other words, when conditional procedure accuracy equality is most likely to be in serious jeopardy, the damage to conditional procedure accuracy will be the greatest. More classification errors will be made; more 1s will be treated as 0s and more 0s will be treated as 1s. A consolation may be that everyone will be equally worse off. 

\subsection{Making Fairness Operational}

It has long been recognized that efforts to make criminal justice decisions more fair must resolve a crucial auxiliary question: equality with respect to what benchmark (Blumstein et al., 1983)? To take an example from today's headlines (Salman, 2016; Corbett-Davies et al., 2017), should the longer prison terms of Black offenders be on the average the same as the shorter prison terms given to White offenders or should the shorter prison terms of White offenders be on the average the same as the longer prison terms given to Black offenders? Perhaps one should split the difference? Fairness by itself is silent on the choice, which would depend on views about the costs and benefits of incarceration in general. All of the proposed corrections for unfairness we have found are agnostic about what the target outcome for fairness should be. If there is a policy preference, it should be built into the algorithm. For instance, if mass-incarceration is the dominant concern, the shorter prison terms of White offenders might be a reasonable fairness goal for both Whites and Blacks.\footnote
{ 
 Zliobaite and Custers (2016) raise related concerns for risk tools derived from conventional linear regression for lending decisions.
}

We have been emphasizing binary outcomes, and the issues are much the same. For example, whose conditional use accuracy should be the policy target? Should the conditional use accuracy for male offenders or female offenders become the conditional use accuracy for all? An apparent solution is to choose as the policy target the higher accuracy. But that ignores the consequences for the false negative and false positive rates. By those measures, an undesirable desirable benchmark might result. The benchmark determination has made tradeoffs more complicated. 

\subsection{Future Work}

Corrections for unfairness combine technical challenges with policy challenges. We have currently no definitive responses to either. Progress will likely come in many small steps beginning with solutions from tractable, highly stylized formulations. One must avoid vague or unjustified claims or rushing these early results into the policy arena. Because there is a large market for solutions, the temptations will be substantial. At the same time, the benchmark is current practice. By that standard, even small steps, imperfect as they may be, can in principle lead to meaningful improvements in criminal justice decisions. They just need to be accurately characterized.

But even these small steps can create downstream difficulties. The training data used for criminal justice algorithms necessarily reflect past practices. Insofar as the algorithms affect criminal justice decisions, existing training data may be compromised. Current decisions are being made differently. It will be important, therefore, for existing algorithmic results to be regularly updated using the most recent training data. Some challenging technical questions follow. For example, is there a role for online learning? How much historical data should be discarded as the training data are revised? Should more recent training data be given more weight in the analysis? 

\section{A Brief Empirical Example of Fairness Tradeoffs With In-Processing}

There are such stark differences between men and women with respect to crime, that cross-gender comparisons allow for relatively simple theoretical discussions of fairness. However, they also convey misleading impressions of the impact of base rates in general. The real world can be more complicated and subtle. To illustrate, we draw on some ongoing work being done for a jurisdiction concerned about racial bias that could result from release decisions at arraignment. The brief discussion to follow will focus on in-processing adjustments for bias. Similar problems can arise for pre-processing and post-processing. 

At a preliminary arraignment, a magistrate must decide whom to release awaiting an offender's next court appearance. One factor considered, required by statute, is an offender's threat to public safety. A forecasting algorithm currently is being developed, using the machine learning procedure random forests, to help in the assessment of risk. We extract a simplified illustration from that work for didactic purposes.
 
The training data are comprised of Black and White individuals who had been arrested and arraigned. As a form of in-processing, random forests was applied separately to Blacks and Whites. Accuracy was first optimized for Whites. Then, the random forests application to the data for Blacks was tuned so that conditional use accuracy was virtually same as for Whites. The tuning was undertaken using stratified sampling as each tree in the forest was grown, stratifying on the outcome classes. This is effectively the same as changing the prior distribution of the response and alters each tree. All of the output can change as a result, which is very different from trying to introduce more fairness when algorithmic output is translated into a decision.

 A very close approximation of conditional use accuracy equality was achieved. Among the many useful predictors were age, prior record, gender, date of the next most recent arrest, and the age at which an offender was first charged as an adult. Race and residence zip code were not included as predictors.\footnote
{
Because of racial residential patterns, zip code can be a strong proxy for race. In this jurisdiction, stakeholders decided that race and zip code should not be included as predictors. Moreover, because of separate analyses for Whites and Blacks, race is a constant within each analysis. 
}

Two outcome classes are used for this illustration: within 21 months of arraignment, an arrest for a crime of violence or no arrest for a crime of violence. We use these two categories because should a crime of violence be predicted at arraignment, an offender would likely be detained. For other kinds of predicted arrests, an offender might well be freed or diverted into a treatment program. A prediction of no arrest probably could readily lead to a release.\footnote
{
Actually, the decision is more complicated because a magistrate must also anticipate whether an offender will report to court when required to do so. There are machine learning forecasts being developed for failures to appear (FTAs), but a discussion of that work is well beyond the scope of this paper.
}
A 21 month follow up may seem inordinately lengthy, but in this jurisdiction, it can take that long for a case to be resolved.\footnote
{
The project is actually using four outcome classes, but a discussion of those results is also well beyond the scope of this paper. They require a paper of their own.
} 

\begin{table}
\caption{Fairness Analysis for Black and White Offenders at Arraignment Using As An Outcome An Absence of Any Subsequent Arrest for A Crime of Violence (13,396 Blacks; 6604 Whites)}
\scriptsize
\begin{center}
\begin{tabular}{|l|c|c|c|c|}
\hline \hline
Race & Base Rate &  Conditional Use Accuracy  & False Negative Rate & False Positive Rate \\
\hline 
{Black} & {.89} & {.93} & {.49} & {.24} \\
{White} & {.94} & {.94} & {.93} & {.02} \\
\hline \hline
\end{tabular}
\end{center}
\label{tab:fair}
\end{table}

Table~\ref{tab:fair} provides the output that can be used to consider the kinds of fairness commonly addressed in the existing criminal justice literature. Success base rates are reported on the far left of the table, separately for Blacks and Whites: .89 and .94 respectively. For both, the vast majority of offenders are not arrested for a violent crime, but Blacks are more likely to be arrested for a crime of violence after a release. It follows that the White re-arrest rate is .06, and the black re-arrest rate is .11, nearly a 2 to 1 difference. 

For this application, we focus on the probability that when the absence of an arrest for a violent crime is forecasted, the forecast is correct. The two different applications of random forests were tuned so that the probabilities are virtually the same: .93 and .94. There is conditional use accuracy equality, which some assert is a necessary feature of fairness.

But as already emphasized, except in very unusual circumstances, there are tradeoffs. Here, the false negative and false positive rates vary dramatically by race. The false negative rate is much higher for Whites so that violent White offenders are more likely than violent Black offenders to be incorrectly classified as nonviolent. The false positive rate is much higher for Blacks so that nonviolent Black offenders are more likely than nonviolent White offenders to be incorrectly classified as violent. Both error rates mistakenly inflate the relative representation of Blacks predicted to be violent. Such differences can support claims of racial injustice. In this application, the tradeoff between two different kinds of fairness has real bite.

One can get another perspective on the source of the different error rates from the ratios of false negatives and false positives. From the cross-tabulation (i.e., confusion table) for Blacks, the ratio of the number of false positives to the number of false negatives is a little more than 4.2. One false negative is traded for 4.2 false positive. From the cross-tabulation for Whites, the ratio of the number false \textit{negatives} to the number of false \textit{positives} is a little more than 3.1. One false positive is traded for 3.1 false negatives. For Blacks, false negatives are especially costly so that the algorithms works to avoid them.  For Whites, false positives are especially costly so that the algorithm works to avoid them. In this instance, the random forest algorithm generates substantial treatment inequality during in-processing while achieving conditional use accuracy equality.

With the modest difference in base rates, the large difference in treatment equality may seem strange. But recall that to arrive conditional use accuracy equality, random forests was applied and tuned separately for Blacks and Whites. For these data, the importance of specific predictors often varied by race. For example, the age at which offenders received their first charge as an adult was a very important predictor for Blacks but not for Whites. In other words, the \textit{structure} of the results was rather different by race. In effect, there was one $h_{B}(L,S)$ for Blacks and another $h_{W}(L,S)$ for Whites, which can help explain the large racial differences in the false negative and false positive rates. With one exception (Joseph et al., 2016), different fitting structures for different protected group categories has to our knowledge not been considered in the technical literature, and it introduces significant fairness complications as well (Zliobaite and Custers, 2016).\footnote
{
There are a number of curious applications of statistical procedures in the Zliobaite and Custers paper (e.g., propensity score matching treating gender like an experimental intervention despite the probability of being female either 1.0 or 0.0). But the concerns about fairness when protected groups are fitted separately are worth a serious read.
} 

In summary, Table~\ref{tab:fair} illustrates well the formal results discussed earlier. There are different kinds of fairness that in practice are incompatible. There is no technical solution without some price being paid. How the tradeoffs should be made is a political decision. 

\section{Conclusions}

In contrast to much of the rhetoric surrounding criminal justice risk assessments, the problems can be subtle, and there are no easy answers. Except in stylized examples, there will be tradeoffs. These are mathematical facts subject to formal proofs (Kleinberg et al., 2016; Chouldechova, 2016). Denying that these tradeoffs exist is not a solution. And in practice, the issues can be even more complicated, as we have just shown. 

Perhaps the most challenging problem in practice for criminal justice risk assessments is that different base rates are endemic across protected group categories. There is, for example, no denying that young men are responsible for the vast majority of violent  crimes. Such a difference can cascade through fairness assessments and lead to difficult tradeoffs. 

Criminal justice decision-makers have begun wrestling with the issues. One has to look no further than the recent ruling by the Wisconsin Supreme Court, which upheld the use of one controversial risk assessment tool (i.e., COMPAS) as one of many factors that can be used in sentencing (State of Wisconsin v. Eric L. Loomis, Case \# 2915AP157-CR). Fairness matters. So does accuracy. 

There are several potential paths forward. First, criminal justice risk assessments have been undertaken in the United States since the 1920s (Burgess, 1926; Borden, 1928). Recent applications of advanced statistical procedures are just a continuation of long term trends that can improve transparency and accuracy, especially compared to decisions made solely by judgment (Berk and Hyatt, 2015). They also can improve fairness.  But categorical endorsements or condemnations serve no one.   

Second, as statistical procedures become more powerful, especially when combined with ``big data,'' the various tradeoffs need to be explicitly represented and available as tuning parameters that can be easily adjusted. Such work is underway, but the technical challenges are substantial. There are conceptual challenges as well, such as arriving at measures of fairness with which tradeoffs can be made. There too, progress is being made.

Third, in the end, it will fall to stakeholders --  not criminologists, not statisticians and not computer scientists -- to determine the tradeoffs. How many unanticipated crimes are worth some specified improvement in conditional use accuracy equality? How large an increase in the false negative rate is worth some specified improvement in conditional use accuracy equality? These are matters of values and law, and ultimately, the political process. They are not matters of science. 

Fourth, whatever the solutions and compromises, they will not come quickly. In the interim, one must be prepared to seriously consider modest improvements in accuracy, transparency, and fairness. One must not forget that current practice is the operational benchmark (Salman et al., 2016). The task is to try to improve that practice.

Finally, one cannot expect any risk assessment tool to reverse centuries of racial injustice or gender inequality. That bar is far too high. But, one can hope to do better. 

\pagebreak

\section*{References}
\begin{description}
\item
Angwin, J, Larson, J., Mattu, S., and Kirchner, L. (2016) ``Machine Bias'' https://www.propublica.org/article/machine-bias-risk-assessments \linebreak-in-criminal-sentencing
\item
Barnett, V. (2012) \textit{Comparative Statistical Inference} New York: Wiley.
\item
Barocas, S., and Selbst, A.D. (2016) ``Big Data's Disparate Impact.'' \textit{California Law Review} 104: 671 -- 732.
\item
Berk, R.A. (2008) ``The Role of Race in Forecasts of Violent Crime.'' \textit{Race and Social Problems} 1: 231 -- 242.
\item
Berk, R.A. (2012) \textit{Criminal Justice Forecasts of Risk: A Machine Learning Approach}. New York: Springer.
\item
Berk, R.A. (2016a) \textit{Statistical Learning from a Regression Perspective,} Second Edition. New York: Springer.
\item
Berk, R.A. (2016b) ``A Primer on Fairness in Criminal Justice Risk Assessments.'' \textit{Criminology} 41(6): 6 -- 9.
\item
Berk, R.A., (2017) ``An Impact Assessment of Machine Learning Risk Forecasts on Parole Board Decisions and Recidivism.'' \textit{Journal of Experimental Criminology}, forthcoming.
\item
Berk, R.A., and de Leeuw, J. (1999) ``An Evaluation of California's Inmate Classification System Using a Generalized Regression Discontinuity Design.'' \textit{Journal of the American Statistical Association} 94(448): 1045 --1052.
\item
Berk, R.A.,  and Bleich, J. (2013) ``Statistical Procedures for Forecasting Criminal Behavior: A Comparative Assessment.'' \textit{Journal of Criminology and Public Policy} 12(3): 513 -- 544.
\item
Berk, R.A., and Hyatt, J. (2015) ``Machine Learning Forecasts of Risk to Inform Sentencing Decisions.'' \textit{The Federal Sentencing Reporter} 27(4): 222 -- 228.
\item
Berk, R.A., and Sorenson, S.B. (2016) ``Forecasting Domestic Violence: A Machine Learning Approach to Help Inform Arraignment Decisions.'' \textit{Journal of Empirical Legal Studies} 31(1): 94 -- 115.
\item
Berk, R.A., Brown, L., Buja, A., George, E., and Zhao, L. (2017) ``Working With Misspecified Regression Models.'' \textit{Journal of Quantitative Criminology}, forthcoming.
\item
Bishop, C.M. (2006) \textit{Pattern Recognition and Machine Learning} New York: Springer.
\item
Blumstein, A., Cohen, J., Martin, S.E., and Tonrey, M.H. (1983) \textit{Research on Sentencing: The Search for Reform, Volume I} Washington, D.C.: National Academy Press.
\item
Borden, H.G. (1928) ``Factors Predicting Parole Success.'' \textit{Journal of the American Institute of Criminal Law and Criminology} 19: 328 -- 336.
\item
Brennan, T., and Oliver, W.L. ``The Emergence of Machine Learning Techniques in Criminology.'' \textit{Criminology and Public Policy} 12(3): 551 -- 562.
\item
Buja, A., Berk, R.A., Brown, L., George, E., Pitkin, E., Traskin, M., Zhao, L., and Zhang, K. (2017) ``Models as Approximations --- A Conspiracy of Random Regressors and Model Violations Against Classical Inference in Regression.'' $imsart-sts  \hspace{.04in} ver. 2015/07/30: Buja\_et\_al\_Conspiracy-v2.tex date: July \hspace{.04in} 23, 2015.$
\item
Burgess, E.~M. (1928) ``Factors Determining Success or Failure on Parole.'' In A.~A. Bruce,  A.~J. Harno, E.~.W Burgess, and E.~W., Landesco (eds.) \textit{The Working of the Indeterminate Sentence Law and the Parole System in Illinois} (pp. 205 -- 249). Springfield, Illinois, State Board of Parole.
\item
Calmon, F.P., Wei, D., Ramamurthy, K.N., Varshney, K.R., (2017) ``Optimizing Data Pre-Processing for Discrimination Prevention,'' arXiv: 1704.03354v1 [stat.ML].
\item
Cohen, A. (2012) ``Wrongful Convictions: A New Exoneration Registry Tests Stubborn Judges.'' \textit{The Atlantic} May, 21.
\textit
Corbett-Davies, S., Pierson, E., Feller, A., Goel, S., and Hug, A. (2017) ``Algorithmic Decision Making and Cost of Fairness.''  asXiv:1701.08230v3 [cs.CY]
\item
Chouldechova, A. (2016) ``Fair Prediction With Disparate Impact:
A Study of Bias in Recidivism Prediction Instruments.'' arXiv:1610.075254v1 [stat.AP]
\item
Crawford, K. (2016) ``Artificial Intelligence's White Guy Problem.'' New York Times, Sunday Review, June 25.
\item
Demuth, S. (2003) ``Racial and Ethnic Differences in Pretrial Release Decisions and Outcomes: A Comparison of Hispanic, Black and White Felony Arrestees.'' \textit{Criminology} 41: 873 --  908.
\item
Dieterich, W., Mendoza, C., Brennan, T. (2016) ``COMPAS Risk Scales: Demonstrating Accuracy Equity and Predictive Parity.'' Northpoint Inc. 
\item
Doleac, J, and Stevenson, M. (2106) ``Are Criminal Justice Risk Assessment Scores Racist?'' Brookings Institute. https://www.brookings.edu/blog/up-front/2016/08/22/are-criminal-risk-assessment-scores-racist/
\item
Dwork, C., Hardt, Y., Pitassi, T., Reingold, O., and Zemel, R. (2012) ``Fairness Through Awareness.'' In \textit{Proceedings of the 3rd Innovations of Theoretical Computer Science}: 214 -- 226.
\item
Feldman, M., Friedler, S., Moeller, J., Scheidegger, C., and Venkatasubrtamanian, S. (2015)  ``Certifying and Removing Disparate Impact.'' In Proceedings of the 21th ACM SIGKDD International Conference on Knowledge Discovery and Data Mining. ACM, 259 -- 268.
\item
Freedman, D.A. (1981) ``Bootstrapping Regression Models.'' \textit{Annals of Statistics} 9(6): 1218 --1228.
\item
Feeley, M. and Simon, J. (1994). ``Actuarial Justice: The Emerging New Criminal Law.'' In D. Nelken (ed.), \textit{The Futures of Criminology} (pp. 173 -- 201). London: Sage Publications.
\item
Friedler, S.A., Scheidegger, C., and Venkatasubramanian, S. (2016) ``On The (Im)possibility of Fairness).'' axXiv1609.07236v1 [cs.CY].
\item
Ferguson, A.G. (2105) ``Big Data and Predictive Reasonable Suspicion.'' \textit{University of Pennsylvania Law Review} 163(2): 339 -- 410.
\item
Hajianm S., and Domingo-Ferrer (2013) ``A Methodology for Direct and Indirect Discrimination Prevention in Data Mining. \textit{IEEE transactions on knowledge and data engineering}, 25(7):1445 -- 1459, 2013.
\item
Harcourt, B.W. (2007) \textit{Against Prediction: Profiling, Policing, and Punishing in an Actuarial Age}. Chicago, University of Chicago Press.
\item
Hardt, M., Price, E., Srebro, N. (2016) ``Equality of Opportunity in Supervised Learning.'' In D.D. Lee, Sugiyama, U.V. Luxburg, I.  Guyon, and R. Garnett (eds.) \textit{Equality of Opportunity in Supervised Learning}. Advances in Neural Information Processing Systems 29: Annual Conference  on Neural Information Processing Systems 2016, December 5-10, 2016, Barcelona, Spain, (pp.3315 -- 3323).
\item
Hastie, T. and Tibshirani R. (1990) \textit{Generalized Additive Models}. London: Chapman Hall. 
\item
Hastie, T., Tibshirani, R., and Friedman, J. (2009) \textit{The elements of Statistical Learning: Data Mining, Inference, and Prediction}, second edition. New York: Springer.
\item
Hamilton, M. (2016) ``Risk-Needs Assessment: Constitutional and Ethical Challenges.'' \textit{American Criminal Law Review} 52(2): 231 -- 292. 
\item
Hyatt, J.M., Chanenson, L. and Bergstrom, M.H. (2011) ``Reform in Motion: The Promise and Profiles of Incorporating Risk Assessments and Cost-Benefit Analysis into Pennsylvania Sentencing.''\textit{Duquesne Law Review} 49(4): 707 -- 749.
\item
Janssan, M., and Kuk, G. (2016) ``The Challenges and Limits of Big Data Algorithms in Technocratic Governance.'' \textit{Government Information quarterly} 33: 371 -- 377.
\item
Johndrow, J.E., and Lum, K. (2017) ``An Algorithm for Removing Sensitive Information: Application to Race-Independent Recidivism Prediction.'' arXIV:1703.049557v1 [Stat.AP].
\item
Johnson, K.D., Foster, D.P. and Stine, R.A. (2016) ``Impartial Predictive Modeling: Ensuring Fairness in Arbitrary Models.'' arXIV:1606.00528v1 [stat.ME]. 
\item
Joseph, M., Kearns, M., Morgenstern, J.H., and Roth, A. (2016) In D.D. Lee, M. Sugiyama, U. V. Luxburg, I. Guyon, and R. Garnett (eds.) \textit{Fairness in Learning: Classic and Contextual Bandits}.Advances in Neural Information Processing Systems 29: Annual Conference on Neural Information Processing Systems 2016, December 5-10, 2016, Barcelona, Spain (pp. 325 -- 333.
\item
Kamiran, F., and Calders, T. (2009) ``Classifying Without Discrimination.'' \textit{2009 2nd International Conference on Computer, Control and Communication}, IC4 2009.
\item
Kamiran, F., and Calders, T. (2012) ``Data Preprocessing Techniques for Classification Without Discrimination.'' \textit{Knowledge Information Systems} 33:1 - 33.
\item
Kamiran, F., Karim, A., and Zhang, X. (2012) ``Decision Theory for Discrimination-Aware Classification.'' IEEE 12th International Conference on Data Mining.
\item
Kamishima, T., Akaho, S., and Sakuma, J. (2011) ``Fairness-aware Learning Through a Regularization Approach.'' Proceedings of the 3rd IEEE International Workshop on Privacy Aspects of Data Mining.
\item
Kearns, M.J. (1994) \textit{An Introduction to Computational Learning Theory} Cambridge: The MIT Press.
\item
Kleinberg, J., Mullainathan, S., and Raghavan, M. (2016) ``Inherent Trade-Offs in Fair Determination of Risk Scores.'' arXiv: 1609.05807v1 [cs.LG].
\item
Kleinberg, J., Lakkaraju, H., Leskovec, J., Ludwig., and Mullainathan S. (2017) ``Human Decisions and Machine Predictions.'' NBER Working paper 23180. National Bureau of Economic Resaerch.
\item
Kroll, J.A., Huey, J., Barocas, S., Felten, E.W., Reidenberg, J.R., Robinson, D.G., and Yu, H. (2017) ``Accountable Algorithms.'' \textit{University of Pennsylvania Law Review}, forthcoming.
\item
Liu, Y.Y., Yang, M., Ramsay, M., Li, X.S., and Cold, J.W. (2011) ``A Comparison of Logistic Regression, Classification and Regression Trees, and Neutral Networks Model in Predicting Violent Re-Offending.'' \textit{Journal of Quantitative Criminology} 27: 547 -- 573.
\item
Messinger, S.L., and Berk, R.A. (1987) ``Dangerous People: A Review of the NAS Report on Career Criminals.'' \textit{Criminology} 25(3): 767 -- 781
\item
National Science and Technology Council (2016) ``Preparing for the Future of Artificial Intelligence.'' Executive of the President, National Science and Technology Council, Committee on Technology.
\item
Pedreschi, D., Ruggieri, S, and Turini, F. (2008) ``Discrimination-Aware Data Mining.'' KDD2008, August 24 -- 27, 2008, Las Vegas, Nevada, USA.
\item
Pew Center of the States, Public Safety Performance Project (2011) ``Risk/Needs Assessment 101: Science Reveals New Tools to Manage Offenders.'' The Pew Center of the States.www.pewcenteronthestates.org/publicsafety.
\item
Powers, D.M.W. (2011) ``Evaluation: From Precision, Recall and F-Measure to ROC, Informedness, Markedness \& Correlation.'' \textit{Journal of Machine Learning Technologies} 2(1): 37 -- 63.
\item
Ridgeway, G. (2013a) ``The Pitfalls of Prediction.'' \textit{NIJ Journal} 271.
\item
Ridgeway, G. (2013b) ``Linking Prediction to Prevention.'' \textit{Criminology and Public Policy} 12(3) 545 -- 550. 
\item
Rhodes, W. (2013) ``Machine Learning Approaches as a Tool for Effective Offender Risk Prediction.'' \textit{Criminology and Public Policy} 12(3) 507 -- 510.
\item
Salman, J., Coz, E,.L., and Johnson, E. (2016) ``Florida's Broken Sentencing System. Sarasota Herald Tribune. http://projects.heraldtribune.com/ bias/sentencing/
\item
Silver, E., \& Chow-Martin, L. (2002) ``A Multiple Models Approach to Assessing Recidivism Risk: Implications for Judicial Decision Making.'' \textit{Criminal Justice and Behavior} 29: 538 -- 569.
\item
Starr. S.B. (2014a) ``Sentencing by the Numbers.'' New York Times op-ed, August 10, 2014.
\item
Starr, S.B. (2014b) ``Evidence-Based Sentencing and The Scientific Rationalization of Discrimination.'' \textit{Stanford Law Review} 66: 803 -- 872.
\item
Thompson, S.K. \textit{Sampling}, third edition. New York: Wiley.
\item
Tonry, M. (2014) ``Legal and Ethical Issues in The Prediction of Recidivism.'' \textit{Federal Sentencing Reporter} 26(3): 167 -- 176.
\item
White, H. (1980) ``Using Least Squares to Approximate Unknown Regression Functions.'' \textit{International Economic Review} 21(1): 149 -- 170. 
\item
Zliobaite, I., and Custers, B. (2016) ``Using Sensitive Personal Data May Be Necessary for Avoiding Discrimination in Data-Driven Decision Models.'' \textit{Artificial Intelligence and the Law} 24(2): 183 -- 201.

\end{description}
\end{document}